# SRE: Semantic Rules Engine For the Industrial Internet-Of-Things Gateways

Charbel El Kaed, *Member, IEEE,* Imran Khan, *Member, IEEE,* Andre Van Den Berg, Hicham Hossayni and Christophe Saint-Marcel

*Abstract*—The Advent of the Internet-of-Things (IoT) paradigm has brought opportunities to solve many real-world problems. Energy management, for example has attracted huge interest from academia, industries, governments and regulatory bodies. It involves collecting energy usage data, analyzing it, and optimizing the energy consumption by applying control strategies. However, in industrial environments, performing such optimization is not trivial. The changes in business rules, process control, and customer requirements make it much more challenging. In this paper, a Semantic Rules Engine (SRE) for industrial gateways is presented that allows implementing dynamic and flexible rule-based control strategies. It is simple, expressive, and allows managing rules on-the-fly without causing any service interruption. Additionally, it can handle semantic queries and provide results by inferring additional knowledge from previously defined concepts in ontologies. SRE has been validated and tested on different hardware platforms and in commercial products. Performance evaluations are also presented to validate its conformance to the customer requirements.

## I. Introduction

The Internet-of-Things (IoT) is expected to interconnect sensors, devices, gateways, and objects on a massive scale [1] to solve many challenges. Such interconnectivity will play an important part to build powerful industrial systems [2] that are more energy efficient with lower costs while contributing to a better environment. In addition, the Semantic Web technology [3] is getting more popular in the development of rich and interactive applications [4]. In industrial environments and automation domains, the Semantic Web has been used to solve data interoperability issues [5], [6] and to provide context aware applications and services [7].

In industrial facilities, black-box devices (gateways or data loggers) are typically used to collect data from variety of devices like; sensors, actuators, machines, plants, processes, and systems. Typically these gateways, sample, collect, and push data to a remote platform for further analysis and may also send notifications for certain events.

In this era of inter-connectivity and optimization, users and managers desire some sort of flexible control over their installations. Due to ever changing business requirements and the nature of IoT applications and services, it has become difficult to preempt the requirements of the users from manufacturer perspective. More dynamic solutions are required that by design offer more flexibility and better control to theses gateways users. Further more, issues like low bandwidth, latency, and security in the industrial domain are additional factors to consider [8]. Hence, easy-to-use solutions located closer to the devices are preferred but issues like device and protocol heterogeneity, multi-vendor solutions, and variety of data models make it a challenging task.

In this paper, we present Semantic Rules Engine (SRE) which consists of two parts: a Rules Engine and a Semantic Engine. The Rules Engine provides a simple and effective way to deploy a control mechanism (as rules) on gateways. These rules are expressed in a simple scripting language and can be modified and uploaded at run-time without disrupting the operation of the gateways. The Semantic Engine provides absolute abstraction from the heterogeneity of devices, protocols, data and any topological changes. It leverages devices meta-data and enables the retrieval of contextual information using semantic queries. Inspired from our previous work [9], we leverage a modular approach with a set of common and domain specific ontologies across our enterprise. For instance, a domain ontology capturing one of our industrial automation contexts, is used to annotate the devices with their contextual information, thereby promoting data interoperability and its understanding to users and applications.

The following are key differentiating factors of SRE. *1)* The rules rely on devices' annotations and are not tied to the devices ID or any other binding. Hence, they are flexible and independent of topology changes and assigning unique identifiers does not affect their execution. *2)* Rules themselves use semantic queries to retrieve contextual information in order to perform actions/eventing functions. Thus, higher-level concepts like, location and measurement type can be used to compose logic in rules to achieve a desired behavior from the gateway. *3)* We have developed a simple reasoning module in SRE to answer on-demand queries. For example, a device's location is in a room *A*, and such room is located on a floor within a building *A*. The reasoning feature will deduce that the device's location is also in building A. This is powerful when aggregating measurements or searching for specific device types. *4)* SRE provides complete life-cycle management of its rules, they can be installed, activated, modified at run-time, deactivated, and uninstalled. *5)* SRE supports cooperation between rules to share functions which are

The authors are with IoT & Digital Transformation, Schneider Electric, Boston One Campus, Andover, MA, 01810, USA and 38TEC, Grenoble, 38000, France. (e-mail: {charbel.el-kaed, imran2.khan}@schneiderelectric.com, andrevandenberg101@gmail.com, {hicham.hossayni, christophe.saint-marcel}@non.schneider-electric.com)

needed by many rules thereby avoiding redundancy. *6)* SRE uses the concept of settable parameters in rules. These parameters can be updated on-the-fly without disturbing the execution of the rules.

The rest of the paper is organized as follows: Section II presents the background, motivating scenario, requirements and the related works. Section III describes the proposed solution while section IV details the implementation and evaluation. Section V outlines lessons learned and future work while Section VI concludes the paper.

## II. BACKGROUND, REQUIREMENTS, AND RELATED WORK

This section presents the background, motivating scenario, requirements and a review of the related works.

### A. Background

Energy consuming devices have evolved from just being able to switch ON or OFF to be more sophisticated for control and monitoring purposes. However, these devices have mainly worked in isolation, only taking into account their own measurements without any sort of intelligence or learning. With the advent of IoT paradigm, the "smart devices" of today are much more advanced than their predecessors. They can communicate with each other using various protocols and can push data of their connected sensors and actuators to a remote cloud platform where it can be analyzed.

However, relying on a cloud platform to remotely control devices is sometimes not desirable for various reasons ranging from the privacy concerns to latency issues and to minimize the downtime or to enable a facility to operate even without the cloud connectivity. More over, once the data has been analyzed in the cloud, normally there is no clear defined mechanism to deploy/modify the behavior of the devices, such as changing the push interval of the data or to select additional data points from a specific set of devices from a particular location.

Thus, there is need to be able to create control strategy (i.e. rules) and execute them on the gateways for more localized control. These rules can range from simply sending an alarm when the temperature in an area reaches a certain threshold, to a more complex rule to control a liquid flow within an underground mining facility [10].

### B. Motivating Scenario and Requirements

In one of our industrial manufacturing facility, an energy monitoring solution has been deployed. The solution consists of an IoT gateway and a set of wireless sensors deployed along the industrial manufacturing machines. The IoT gateway collects data from the sensors and pushes it to the remote cloud platform for visualization and analysis, as shown in Fig. 1.

In our continuous efforts to maintain green facilities, an internal energy audit concluded the following: most of the energy waste is occurring in the semi-automated industrial process. Energy can be saved by engaging our local personnel working on the industrial process. One simple solution is to install light indicators, along the industrial process, to inform the local personnel about situations like energy wastage (Orange light ON), process fault (Red light ON) or normal operation (Green light ON). The audit team along with the facility manager proposed a set of rules to create this solution.

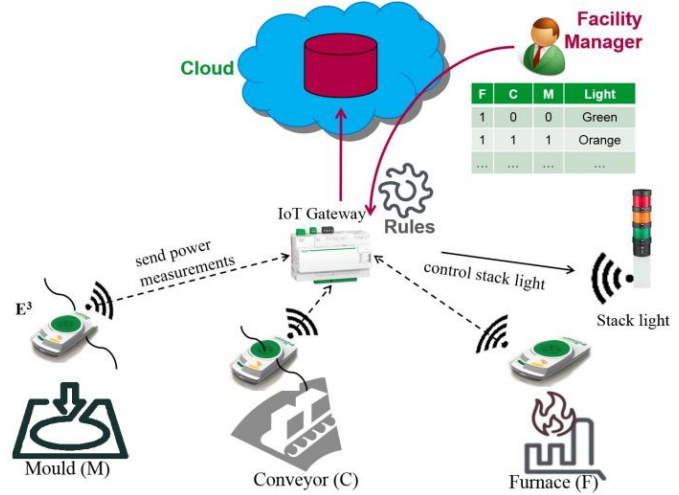

Fig. 1. Our Industrial Facility Use Case

For example, in Fig. 1, when the conveyor is ON while the mold and the furnace are OFF, energy is being wasted. However, for preheating purposes, it is acceptable for a short period of time for the furnace alone to be ON.

From this simple scenario, it is clear that the facility manager needs a solution that gives her more flexibility to tackle new situations that were not envisioned during the initial setup. The solution needs to be dynamic, flexible and should not involve any reprogramming or hard reset of the devices. It should also be usable by non-IT experts.

There are two possible solutions : One is to use a tailored solution to solve one problem only and when new requirements arise, reprogram everything again. This option does not scale well and is not suitable to solve new problems or implement new business rules efficiently. The second option is to have a more flexible solution that is easier to implement, does not require reprogramming and supports quick implementation of new business rules. Thus, the solution requires the following characteristics: 1) Expressing requirements and conditions in an explicit manner to control the devices and to send alerts. 2) Deployment strategy of these requirements. 3) Efficient execution of these requirements. 4) Support semantics to query status and details of the devices by location, type, connectivity, and other concepts. 5) Additionally, the solution should be suitable for different types of devices (gateways, edge, fog nodes, and Raspberry Pi), easily integrate-able, modular in design to implement new features, and be independent of device types, protocols or application domain constraints.

In literature, we found several Rules Engines and Semantic Engines, we discuss some of them in the following subsection.

*C. Related Work*

The work in [11] proposes LINC, a coordination programming environment to execute Rules efficiently. LINC uses different Linda-like tuple-spaces. The global tuple space is split into "bags". Each bag can either encapsulates a database, service (SOAP or RPC), event system or hardware like a sensor or an actuator. A coordination language is used to specify Rules with statements and has operation types get, put and read to interact with the "bags" and identify their content using keys or identifiers. Dependence on identifiers and keys makes LINC unsuitable for dynamic environment because rules will need updates as well. This is not the case in SRE, which remains immune to such changes thanks to the use of semantic queries. Also, compared to SRE, LINC does not provide any query mechanism except the keys to retrieve objects from bags. Therefore it is not possible to look up various things on the gateway based on their contextual information such as their measurements types and/or physical locations. Moreover, a simple rule example as described in [11] does not seem trivial for a developer compared to Lua in SRE following an EventCondition-Action paradigm. It is also not clear if LINC uses higher-level of abstraction for Rules similar to SRE.

Authors in [12] propose a Rules Engine to efficiently handle the high data throughput and large set of rules in the context of smart building systems. Due to high frequency of data, it is difficult to apply corresponding Rules from a large set, for example to handle urgent events. The sensor and actuator data are used to find atomic events then rules are used to construct the minimum prefect hash table to filter meaningless atomic events. Using a feedback mechanism the rule matching overhead is further reduced. The main strengths of this approach are scalability and rules conflict management with the help of the user. However, in their approach, a rule is described in an XML data structure limiting the expressiveness of the rule, e.g., it is not clear how a loop is handled or whether atomic conditions with logical operations can be expressed to trigger an action. In addition, each rule is tied to a sensor identifier along with a room and a user identifier making it very rigid when a device is replaced or its physical location changes. SRE does not have these drawbacks because it relies on the Semantic Query Engine to handle dynamic environments and the Lua language constructs for better expressiveness.

There are several other works that address the same problem using approaches such as programming models like [13] and [14], middleware like [15], coordination languages like [16] and [17] and the solutions based on tuplespace like [18], [19], [20] and [21]. Most of these works provide complex solutions that are tailored for a particular scenario or are difficult to use in medium to small gateways. Also some of these works are prior to the IoT paradigm and do not consider the challenges it introduces.

Other existing Rules Engines such as [22], [23] are designed to run with an underlying Java Virtual Machine with a clear dependency on java packages as shown in the import declaration of a rule. In fact, our early investigation started with drools [22], however, due to its rigid dependency on Java and its packages, we opted for a Lua based approach which is light weight and portable even on smaller footprint gateways.

On the Semantic Web side, the authors in [24] discuss a unified semantic engine for IoT and Smart Cities to tackle data interoperability and scalability among others. However, their work does not provide any details regarding the implementation or technologies used to address the identified issues. A Semantic gateway is proposed in [25] to tackle the industrial field service use-case where service engineers rely on a plethora of tools to track and solve the issues of the products which may have been installed decades ago. The semantic aspect in this work only deals with the mapping of the SOAP bindings to the networking platform. A gateway architecture is presented in [26] that abstracts the network services and translates network services to a standard DPWS interface but there is no notion of queries or rules.

The authors in [27], [28] present a semantic smart gateway framework to achieve device interoperability. Features such as 'on-the-fly' ontology learning and ontology alignment are provided. Similarly, our previous work[6] proposes a semiautomatic approach to dynamically generate and install proxies on the gateway from ontology alignments to achieve semantic interoperability. However, these related approaches do not fulfill our requirements of query and control using rules. In our approach, instead of relying on ontology alignment, we opted for the commissioning phase described in our previous work SQenIoT [9] to semantically annotate data sources from ontologies when the gateway is commissioned.

### III. PROPOSED SOLUTION

By carefully considering requirements, constraints and viewpoints of the stakeholders we concluded that our proposed solution will have the following features: *Flexibility*: design rules to express, deploy, and execute user requirements quickly with time and cost savings and avoiding software development life-cycle for every requirement. *Deployment*: install rules on the gateways either locally or remotely through a cloud application. *Interoperability*: leverage the ontology concepts to provide consistent, reusable, and shareable view about the data. *Semantic Query*: rely on a natural language-like grammar to retrieve information from the gateway based on annotated data liberating rules from static unique identifiers. *Modularity*: to easily integrate new functionalities and to scale. Finally, the solution should provide dedicated rules execution environment.

After a careful study of the state-of-the-art and available open source rules engines like [22], [23], we decided to design the Semantic Rules Engine (SRE) to fulfill the required features mentioned earlier. The SRE comprises of two parts: the Rules Engine and an existing work Semantic Engine [9]. The overall design architecture of SRE is shown in Fig. 2. The components are detailed in the following subsections.

*A. Rules Engine*

The Rules Engine provides gateways with the decentralized intelligence and the ability to control or query the connected devices. Figure 3 shows the components of the Rules Engine. A Rules Manager interacts with remote applications, receives and manages the rule files. Once started, the Rules Manager loads the execution environment used to execute the rules. An execution environment provides a set of functions, for example to handle subscriptions, timers, and interactions with other components like SQenIoT and the Cloud Connectivity Agent. Once the execution environment is setup, the Rules Manager handles the life-cycle of the rules (e.g. install, start, stop or delete). Each rule is executed in isolation to avoid conflicts with other rules and resources access issues.

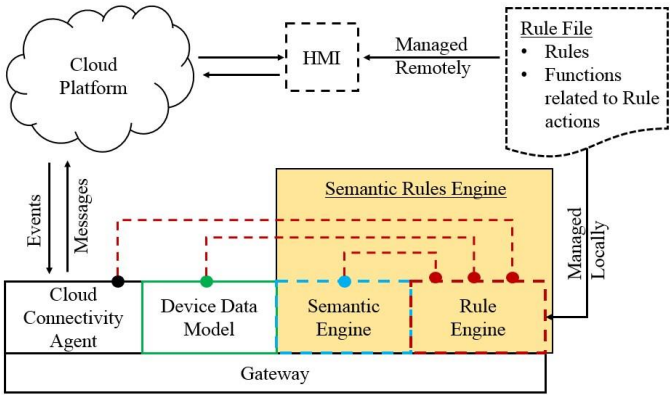

Fig. 2. Design Architecture of Semantic Rules Engine

*B. Semantic Engine: SQenIoT*

To provide the rules with the ability to integrate semantic queries, we reused existing work SQenIoT [9]. An API is provided with SQenIoT which makes it possible for SQenIoT to access the device meta-data. The values of the meta-data tags can be injected during the commissioning phase manually or automatically as described in [9]. By combining Rules Engine with SQenIoT, rules are able to execute semantic queries and use the returned results. For example, a rule can delegate the aggregation of temperature of a given floor to SQenIoT through a query. Then, the returned results will be used in another query to trigger an action or an event.

The formulated query will rely on annotated data in the gateway such as location (i.e. a floor) and measurement (i.e. temperature) tags as means to get the list of temperature sensors to get the average temperature of a given floor. Once received by SQenIoT, it gets the values of the temperature sensors of that particular floor only and returns the average temperature values. The Rules Engine can then compare the result with a given threshold and execute the action (a notification, an alarm, or an actuation) according to the logic specified in the rule. This semantic query feature decouples the rule from any identifier of a sensor or measure and is capable of handling changes in the topology (e.g. device joining/leaving/being replaced).

The interaction between the Rules Engine and SQenIoT is a major differentiating factor from other solutions. In fact, existing Rules Engines rely on a unique identifier to address a sensor, actuator, or measurement. Since the identifier is hardcoded in the rule, it requires manual update when, for example, a faulty sensor is replaced by a new one. On the other hand, by relying on contextual information and meta-data (tags), the rule will still be valid even when the sensor is replaced since the meta-data will remain the same for the new sensor.

Overall, with this capability, SRE offers a flexible solution where the same rules can be reused in different products and solutions as long as higher-level concepts (like location) are reused. This provides time and cost savings as well as same level of functionality across different domains and platforms.

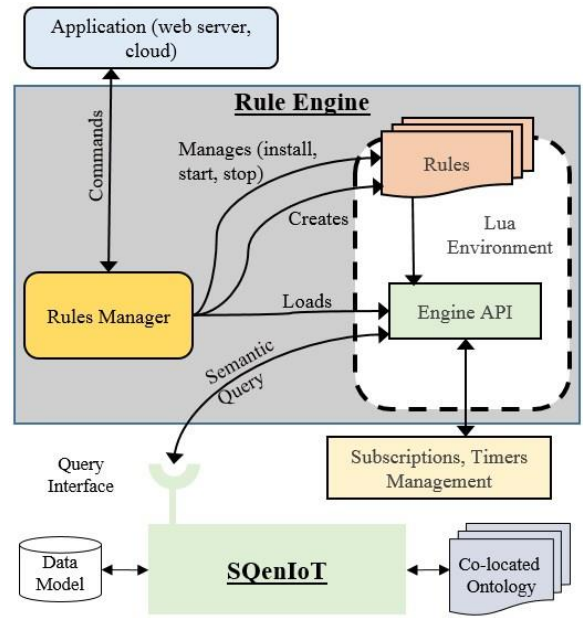

Fig. 3. SRE Implementation Architecture

*C. Rule Design*

The users' requirements, regarding control or information are expressed in the form of rules. Therefore the rules are designed to enable the following:

1) Event-Condition-Action (ECA) paradigm on the resources (i.e. devices/sensors/actuators) and their capabilities (measurement, points).
2) Publish/Subscribe mechanism on a resource or measurement with a related condition.
3) Support life-cycle management process (install, start, stop, and uninstall).
4) Orchestration and cooperation between rules which allow delegation and reuse.

These features are elaborated as follows:

*1) ECA Design:* The condition part of a rule is designed to be flexible. It takes only one resource or capability into account. To evaluate multiple resources and their capabilities, a rule can have multiple conditions combined by logical operators. The following syntax enables to look up "SensorA" and its measurement value "Temperature":

[SensorA] Temperature

Using this definition, multiple conditions can be combined together as follows and are also shown in Fig.4

[MultiSensorA]Temp > 25° C AND
[DoorSensorB]isOpen == True

The supported logical operators are AND and OR, whereas the allowed comparators are: $=, <, ==, \leq, \geq, <>, >$.

The following types of conditions are defined to evaluate resources and their capabilities.

A) *Evaluator*: Evaluates the measurement of a resource and its capability: [SensorA]Temp>25°C It is useful when a capability reaches a threshold or is equal (or not) to a specific value.

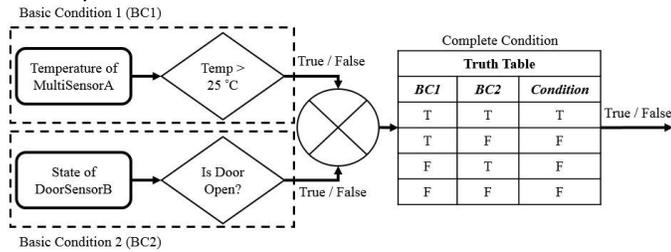

Fig. 4. Example of Condition Consisting of Two Basic Condition Blocks

B) *Existence*: This condition is used to check if a resource exist, for example, in a dynamic context, appearance/disappearance of a device. @exist[SensorA] == True

C) *EventOnChange*: A granular subscription on the change of state/value of a capability (e.g. window is open or close) is of interest.

@change[DoorSensor1]State == True

D) *Increment/Decrement*: Enables the subscription on the increase/decrease of the capability value. @incr[SensorA]Temp == True (@decr)

The *Existence*, *EventOnChange*, and *Increment/Decrement* operations are coupled with a callback function. The logical operators *AND* and *OR* can be combined to specify powerful conditions in the rules. By relying on SQenIoT, users do not need to know the name of the resource and capability they are interested in, instead they can use higher-level concepts like location, device type, and physical quantity as described in [9].

*2) Publish/Subscribe Mechanism:* There are two mechanisms to express publish/subscribe requests. The first one is identifier dependent, which means that the rule must contain the unique identifier of the sensor and its resource, for example, SRE.subscribe(@exist[Sensor11]

== True, HandleDynamicSensorCallback)[1]. The second mechanism goes through SQenIoT with a semantic query based on the sensor metadata instead of its unique identifier, for example, SRE.query("subscribe catalog:sensor and location:Room1", HandleDynamicSensorCallback).

*3) Rule Life-cycle Design:* One of the requirements for the rules is to be able to manage them easily. For example, a rule that lowers the temperature of a hot water geyser might only be started on weekends and stopped on weekdays. We designed life-cycle management of rules, as shown in Fig.5, to easily manage them using a Graphical User Interface (GUI).

*4) Orchestration and Cooperation:* To avoid rule conflicts and to enable reuse, orchestration is primordial. Consider for example, an eventing function which sends an notification with a description to a remote monitoring center, such function can be implemented once and reused by different rules. Therefore, SRE enables rules cooperation and orchestration.

*5) Rule Persistence:* Rules are persisted on the device along with their states (installed, started, stopped) and the recent values of their settable parameters. When a rule is created, it is first encoded in Base64 then the resulting JSON file is digitally signed (for security) and compressed as ZIP file. SRE receives this ZIP file and validates its signature. If valid, the ZIP file is stored locally and the rule is extracted and loaded in the SRE for installation. Additionally, SRE saves the state of the rules locally in a small database with details such as name of the rules, list of settable parameters, and their most recent values. If device reboots, SRE first validates the local ZIP files, extracts the rules and, by using the details in the database, restores them to their last execution state along with the last stored values of the parameters.

Finally, although it is not the focus of this paper, we did consider the security aspects to allow safe deployment of rules in our gateways for execution. Only rules, signed with corporate PKI, are executed by SRE. Every received rule is first validated to check its origin and authenticity and then installed in the gateway, otherwise it is discarded.

### IV. IMPLEMENTATION & EVALUATION

In this Section, we first present the SRE implementation choices and then the setup. Later, performance metrics and the results are discussed. The section concludes with the presentation of real use-cases where SRE is being used.

*A. Implementation Choices*

*1) Rules Engine:* is expected to run on embedded devices as well as resourceful devices. There were quite a few choices like JESS, JBoss Drools [22], Lua and OpenRules. However, Lua [29] was selected due to the following factors: 1) Developed for embedded devices but it is also usable for resourceful devices. 2)

---
[1] https://github.com/InnovationSE/SRE/blob/master/TII/LuaAlertEx.lua

Open source, has a strong user base, and its interpreters are available in several languages including Java and C for integration within different platforms. 3) Easy to use syntax. 4) Expressiveness of basic and complex rules. 5) Multiple execution environments to run rules independently. The Lua interpreter can exist in a separate process on the gateway and therefore has less interference with the main framework and can be shutdown and restarted independently. The design of Rules Engine is shown in Fig. 3. A rule is basically a Lua script executed by the Lua interpreter. Both Java and C version of Rules Engine were developed for different hardware platforms offering same functionality. Using the Engine API, rules can interact with functions to handle subscriptions and timers. The execution environment is created by the Lua interpreter and ensures that multiple rules co-exist without any issues. Also, rules can call Java methods or C functions for more complex operations.

An example of a Lua rule is shown in Listing 1, it regulates the luminosity levels of the lights in a given site based on the average luminosity level. Lines 10, 11, 15, and 16 of Listing 1 demonstrates the expressivity of SRE through semantic queries. Lines 10-11 shows a semantic query which delegates to SQenIoT, the average luminosity value from the Sensors located in Site1. This query makes it very easy to retrieve the aggregated value from the sensors without a reference to any fixed identifier or other bindings. In addition, the inference is requested through the @ symbol in the query. To calculate the inference, SQenIoT, relies on the annotated data and the location ontology created during the commissioning phase which contains concepts and relationships between Site1 and all locations in it (e.g. floors, rooms). The Sensors are not directly tagged with location:Site1, they were annotated with their locations (mainly rooms and open spaces). SQenIoT relies on the ontology to infer and retrieve the sensors located in the site. This feature is a major differentiating factor of SRE from other solutions.

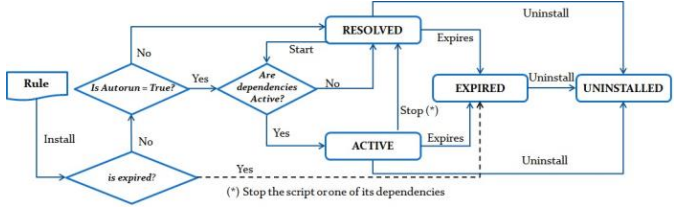

Fig. 5. Lifecycle of a Rule

If the average luminosity is below a certain threshold, then the light actuators are queried (lines 15-16) and a new value is set in lines 18-22 of the Listing. Another interesting feature is on line 7 which allows a rule to get threshold value dynamically. A new threshold value can be provided on the fly and it will be taken into the account by the rule in its next execution cycle. Several semantic queries can be combined together to compose complex rules. Lua also supports userdefined functions so complex computations can be performed in these functions for more advanced control features.

*2) SQenIoT:* The implementation is shown in Fig. 3. Like the Rules Engine, it is also developed in both Java and C languages. It exposes a Query Interface through which rules can send semantic queries. The necessary functions are provided by the Engine API module. For semantic support SQenIoT uses open source libraries, namely OWL API[2] and Redland RDF Libraries[3] in Java and C version respectively.

*B. Implementation Setup*

SRE Java version is deployed on a commercial off-theshelf industrial Gateway which has Modbus[4], Ethernet, WiFi, ZigBee, and GPRS interfaces to connect to devices and a remote cloud platform. The gateway uses Linux OS with IBM J9 [30] as Java Virtual Machine (JVM) on top. J9 is certified for embedded systems in industrial usage. The ProSyst[5] implementation of Open Service Gateway initiative (OSGi) framework[6] is used over J9 for modular design. SRE is implemented as one of the OSGi component.

SRE C version is implemented for our in-house hardware platform based on dual core Cortex A9 chip clocked at 900Mhz with 1GB RAM. It also runs Linux[7] for embedded systems. The connectivity options include, ZigBee, Ethernet, Bluetooth and Wi-Fi. SRE is developed as a POSIX compliant library that can be imported by client applications.

```
1   function LightControl.init()
2   engine.timer("LightControl", "Control", 500,2000,-1)
3   defaultThreshold = 600
4   end

6   function LightControl.Control()
7   threshold = engine.getRuleSetting("LightControl","threshold") 8 if(~threshold)
9       threshold = defaultThreshold
10      averageLuminosityOfSite1_query =
11      "Avg variable usage:LuminositySensor and @loc:Site1"
12      averageLuminosityOfSite1_result
13      = engine.query(averageLuminosityOfSite1_query)
14      if (averageLuminosityOfSite1_result < threshold)
15      light_actuators_devices =
16      "Search Device usage:LightActuator and @loc:Site1"
17      lightActuators = engine.query(light_actuators_devices)
18      for i = 1, #lightActuators, 1 do
19      lux = engine.getCapability(sensors[i],"LuminositySetPoint")
20      differenceToSet = threshold - lux.value
21      if(differenceToSet > 0)
22      engine.setValue(lux, lux.value+differenceToSet)
23      end
24      end
25      end ...
```

---

[2] http://owlapi.sourceforge.net
[3] http://librdf.org
[4] http://www.modbus.org
[5] http://www.prosyst.com
[6] https://www.osgi.org
[7] http://www.windriver.com/products/linux

Listing 1: Lua Rule Example

*C. Performance Metrics*

The performance of SRE is assessed in terms of Rule Execution Time (RET), Rule Latency Time (RLT) & Rule Memory Usage (RMU). RET and RLT are measured in milliseconds.

RET measures the time needed to execute rules. It includes the time needed to parse the rule condition, its evaluation, and the execution of an action. To evaluate RET multiple sets with different number of rules are defined. Each rule defines a condition on 5 smart devices. The condition part results in a true evaluation each time to ensure that the action (i.e. toggle the state of the actuator and send message to cloud) is executed.

RLT measures the average latency between the occurrence of an event and the execution of the corresponding rule. For this we simulated occupant comfort scenario in three multi-story residential buildings to control the heater and air conditioner based on temperature. Each building has ten floors, which have three rooms each. Each room has a temperature sensor. In order to test the robustness of SRE we stress-tested it using a Lua script[8] to control the heater and air conditioner in each room when it's too cold or too hot respectively. SRE evaluates and executes rules in a sequential manner so it is important to determine how much latency is introduced when multiple rules are enabled. We define different sets of rules where each rule monitors a temperature sensor. Then, periodically and simultaneously, multiple events are generated on all temperature sensors to activate corresponding rules.

RMU measures the amount of memory needed to manage certain number of rules, it is measured in Kilobytes.

*D. Results*

For RET, recall that the action part of the rules consists of changing the value of a device and sending a message to the cloud. Figure 6 shows the results of the RET evaluation, it shows that only 12ms are necessary to evaluate and execute 100 rules, while 56ms are needed to evaluate 500 rules. We can conclude that on average the evaluation and execution of each rule takes about 0.12 ms. RET shows that the Rules Engine is suitable to execute large number of rules, however in real situations, executing 100s of rules will not be common.

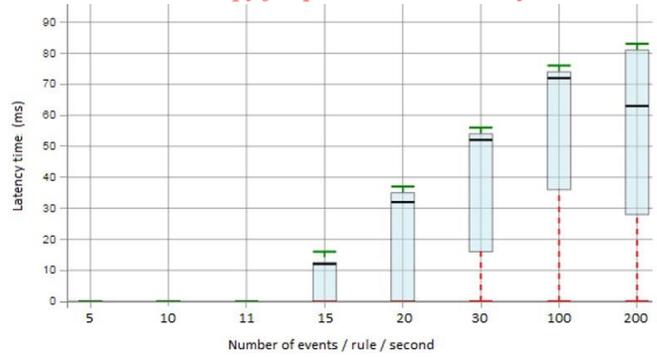

Fig. 7. Rules Latency Time Evaluation

Figure 7 shows the results of RLT evaluation. It shows the averages of latency time of different scenarios. Each scenario consists of the execution during 2 minutes of a specific number of rules between (1, 10, 30, 50, 70, 90, 100, 200, 300, 400), and the generation of a number of events for each rule among (5, 10, 11, 15, 20, 30, 100, 200) events per second. It is found that for all sets of rules, the delay is negligible when the number of events/second are less than or equal to 11. After this latency becomes noticeable and exceeds 80ms for 200 events.

Figure 8 shows RMU, i.e. amount of memory needed by SRE to manage different number of rules. We found that even for large number of active rules, the amount of memory remains acceptable for embedded systems (less than 2400KB).

We find that SRE is suitable for embedded as well as resourceful devices. Its code size is also manageable, as Java version is ≈1.3MB and C version is less than 1MB in size.

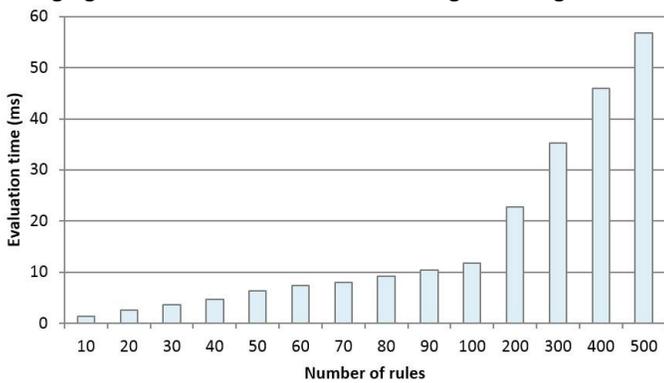

Fig. 6. Rules Execution Time

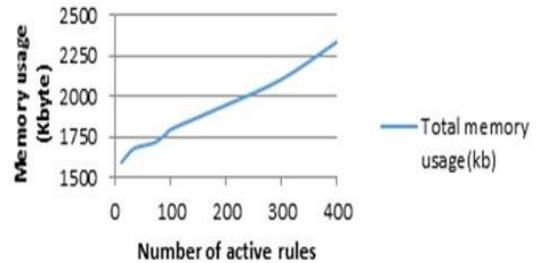

Fig. 8. Memory Usage

---

[8] https://github.com/InnovationSE/SRE/blob/master/TII/SreEvalRule.lua

*E. SRE Deployment in Real Industrial Environments*

We have also tested SRE to solve our real business use cases. The Java version has been tested in an industrial environment to monitor the production lines within a factory to ensure safety of the production lines. The setup is shown in Fig. 9. The rules in SRE receive measurements from Zigbee Green Power modules and then generate alerts, as blinking light, to alert the on-site supervisor. The same SRE version is used in a different facility where a rule is used to detect the tripping event of a circuit breaker, read specific values from Modbus registers to determine the causes of the trip and send this data to the remote monitoring platform without any human intervention. In another deployment, SRE Java C version resides in a gateway and uses rules to, for example, stop sending the data to the cloud when the luminosity is zero in order to save cost on the 3G network.

In terms of added value of SRE, we can envision an application that receives data from multiple SRE deployments and then displays comparison/bench-marking of energy, occupancy, efficiency, and other dimensions. Different widgets can be used to send semantic queries to SRE and display the results in an appropriate manner.

Similarly, an analytics application can get aggregated data from SRE by sending a semantic query. As part of machine learning process, semantic queries can be used to subscribe to the data from machines for a specific duration which can be then fed into an algorithm to learn about the machine failures with real values. In fact, with SRE rules, a complex machine learning strategy can be developed and executed at the Edge without depending on input from the Cloud.

We can see that the rule-based approach in SRE makes it much easier to handle different requirements. There is no need to code and deploy a new solution where requirements change. The rules can be easily managed thanks to our rule life-cycle approach. Our experience shows that SRE is a general purpose solution to solve different problems in different domains.

## V. Lessons Learned and Future Work

During this work we learned several lessons and plan to tackle them in future works. The first lesson is that when there are multiple rules executing concurrently, conflicts may occur. However, detecting conflicts in rules and solving them is not trivial. The works such as [12], and BZR language [31] and [32] can be explored to find a solution. The second lesson is about using properly aligned ontologies, for the meta-data of the devices. This raises ontology governance issue in large enterprises. There is a tremendous scope of using ontologies in industrial context. We can find early initiative like SAREF ontology [33] and its proposed extensions, Haystack Tagging Ontology [34] and IEEE Ontologies for Robotics and Automation [35].

The third lesson is about efficient inference engines usable on gateways and edge devices. Localized inference capabilities can be particularly useful for emerging trends like local analytics [36]. The fourth and final lesson is the need for a userfriendly way to write rules and semantic queries. Developers can easily use Lua to write rules but we still need to abstract them with GUI to make it easier for non-technical users. The solution should consider QoE and the human experience when interacting with technology [37].

## VI. Conclusion

This paper presented SRE with control and querying capability on top of existing gateways and data logger devices. The Rules Engine allows different stakeholders to define rules to control and monitor devices. Rule-based solutions bring flexibility to quickly address business requirements without going through the complex implementation and installation process. There is a complete life-cycle management of rules as well as corporation between rules to enable reuse of functionality. The Semantic Engine, SQenIoT, allows access to device data using semantic tags. The cooperation between the Rules Engine and SQenIoT is what sets this work apart from existing solutions. The rules are defined using semantic tags so that they do refer to higher level concepts and not specific devices. The advantage of this is that rules remain valid even when devices are added or removed.

SRE solved simple yet important requirements related to cost and time savings. Instead of implementing custom functions for each of our end clients by going through typical software development process, SRE makes our industrial gateways customizable with little effort using rules and semantic queries. The work described in this paper has been implemented, tested and validated in real industrial settings. Several work items have been identified and will be used to extend this work.

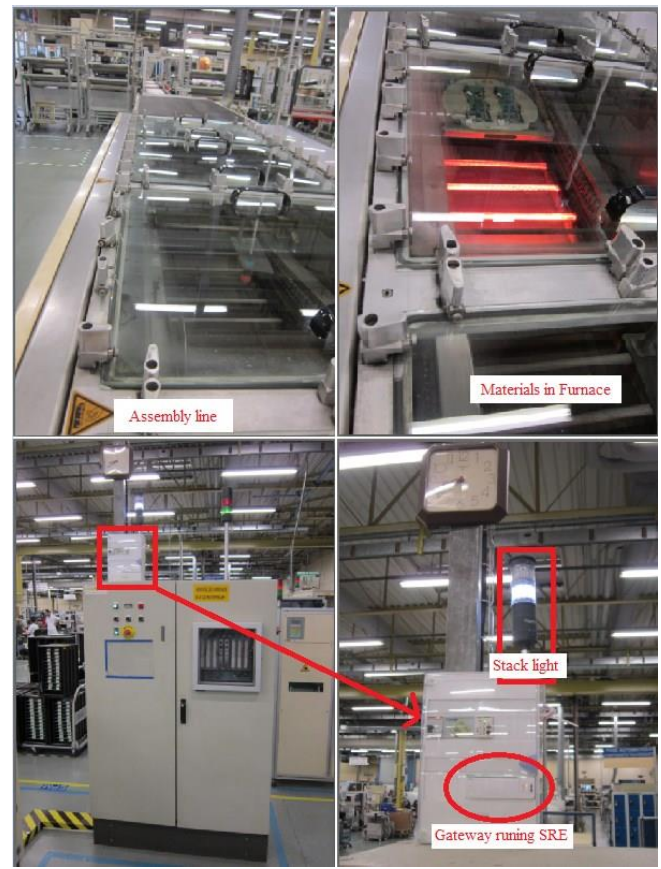

Fig. 9. SRE Deployment in Industrial Environment